\documentclass[conference]{IEEEtran}
\IEEEoverridecommandlockouts
\usepackage{cite}
\usepackage{amsmath,amssymb,amsfonts}
\usepackage{algorithmic}
\usepackage{graphicx}
\usepackage{textcomp}
\usepackage{xcolor}
\usepackage{array,multirow,graphicx}
\usepackage{adjustbox}
\usepackage{amsmath}
\usepackage{comment}
\usepackage{url}
\usepackage{graphicx}
\usepackage{pgfplots}
\usetikzlibrary{pgfplots.groupplots}

\def\BibTeX{{\rm B\kern-.05em{\sc i\kern-.025em b}\kern-.08em
    T\kern-.1667em\lower.7ex\hbox{E}\kern-.125emX}}
\begin{document}

\title{Unsupervised Network Intrusion Detection System for AVTP in Automotive Ethernet Networks}

\author{Natasha Alkhatib, Maria Mushtaq, Hadi Ghauch, Jean-Luc Danger 
\\
\textit{Télécom Paris, IP Paris, Palaiseau, France}
\\
{\fontfamily{qcr}\selectfont
\small{\{natasha.alkhatib, maria.mushtaq, hadi.ghauch, jean-luc.danger\}@telecom-paris.fr}
}
}

\maketitle

\begin{abstract}
Network Intrusion Detection Systems (NIDSs) are widely regarded as efficient tools for securing in-vehicle networks against diverse cyberattacks. However, since cyberattacks are always evolving, signature-based intrusion detection systems are no longer adopted. An alternative solution can be the deployment of deep learning based intrusion detection system which play an important role in detecting unknown attack patterns in network traffic. Hence, in this paper, we compare the performance of different unsupervised deep and machine learning based anomaly detection algorithms, for real-time detection of anomalies on the Audio Video Transport Protocol (AVTP), an application layer protocol implemented in the recent Automotive Ethernet based in-vehicle network. The numerical results, conducted on the recently published "Automotive Ethernet Intrusion Dataset", show that deep learning models significantly outperfom other state-of-the art traditional anomaly detection models in machine learning under different experimental settings.
\end{abstract}


\section{Introduction}
Since the advent of powerful electronic components such as sensors and actuators as well as a robust in-vehicle infrastructure for efficient data exchange between them, driving has become safer (i.e. 360- degree surround view parking assistance, and collision avoidance systems) \cite{ADAS_safety} and more pleasant (i.e. infotainment features)\cite{Infotainment1} \cite{Infotainment2} during the last several decades. Ethernet, a flexible and scalable networking technology in communication systems, is recently standardized and adopted for in-vehicle communication \cite{AutomotiveEthernet1}\cite{AutomotiveEthernet2} between different Electronic Component Units (ECU). In fact, it fulfills basic automotive requirements which existing in-vehicle protocols LIN, CAN, and FlexRay are not designed to cover, including reduced connectivity costs, cabling weight and support for high data bandwidth. 

To ensure low-latency and high-quality transmission of time-critical and prioritized streaming data for high-end infotainment and ADAS systems, the IEEE 1722 audio-video transport protocol (AVTP)\cite{1722} is adopted. In fact, AVTP specifies a protocol for audio, video, and control data transportation on a Time-Sensitive Networking (TSN) capable network \cite{AVB}. As a result, we believe that AVTP protocol will be a critical protocol for Automotive Ethernet-based in-vehicle network in motor vehicles.

Despite the advantages of Automotive Ethernet, the drive toward connectivity has significantly expanded the attack surfaces of automobiles, making Automotive Ethernet-based in-vehicle networks increasingly susceptible to cyberattacks, posing significant security and safety issues \cite{cyberattack}. In fact, Automotive Ethernet can be attacked by exploiting its vulnerabilities \cite{ethernetsecurity}\cite{aesecurity}. These security breaches can affect protocols working on top of it, including AVTP protocol, and might therefore lead to the interruption of critical media streams.

To address this, intrusion detection systems (IDS) should be used in addition to specific security measures as an extra layer of protection. These systems can be classified based on their analyzed activity (i.e., monitoring a network or a host activity  logs)  and  their  detection  approach  (i.e.,  signature-based  or  anomaly-based  detection). Deep learning models, usually referred to as anomaly-based intrusion detection techniques, are in general neural network models with a large number of hidden layers. These models can learn extremely complicated non-linear functions, and their hierarchical layer structure allows them to acquire meaningful feature representations from incoming data. Researchers have explored deep learning techniques for in-vehicle intrusion detection on Controller Area Network (CAN) bus protocol since 2015 \cite{Kang} \cite{leblanc}. However, due to the lack of relevant and public datasets, few studies have been conducted to study the intrusion detection performance of deep learning based IDS for automotive systems using Automotive Ethernet-based network. Among them, Alkhatib et al. \cite{natasha} proposed a deep learning-based sequential model for offline intrusion detection on Scalable Service-Oriented Middleware over IP (SOME/IP) application layer protocol on top of Automotive Ethernet. Moreover, Jeong et al \cite{avtp} presented an intrusion detection method for detecting audio-video transport protocol (AVTP) stream injection attacks in Automotive Ethernet-based networks.

In this paper, we compare the performance of different deep and machine learning based intrusion detection systems for real-time detection of anomalies on the AVTP protocol. Regarding deep learning based models, we leverage different types of autoencoders which reconstructs a sequence of exchanged AVTP packets over the in-vehicle network. Anomalies in AVTP packet stream, which may lead to critical interruption of media streams, are therefore detected by computing the corresponding reconstruction error. These models are compared with other state-of-the-art anomaly detection models such as One-class SVM (OCSVM), Local Outlier Factor (LOF), and Isolation Forest. The numerical results, conducted on the recently published "Automotive Ethernet Intrusion Dataset", show that deep learning based models outperform other baselines under different experimental settings.

The main contributions of this paper are as follows:
\begin{itemize}
    \item We compare the performance of different unsupervised anomaly detection method to detect unknown cyberattacks in real-time on AVTP protocol used in Automotive Ethernet-based in-vehicle network for media streaming. 
    \item We evaluate their performance by using the recently published "Automotive Ethernet Intrusion Detection" dataset \cite{avtp_dataset} and which contains replay attacks.
\end{itemize}

Towards this end, our paper is organized into six sections. In Section \ref{s2}, we present an overview of media stream transportation using AVTP network protocol. In Section \ref{s3}, we present an overview of the considered AVTP dataset, the covered threat model along with the engendered cyberattacks. Section \ref{s4} discusses the detection of in-vehicle network anomalies using unsupervised anomaly detection algorithms. In Section \ref{s5}, we present our evaluation metrics. We discuss our experimental results in Section \ref{s6}. The limitations of our work are presented in Section \ref{s7}. Finally, we conclude our paper with future work direction.

\section{Transmission of Media Streams using AVTP}
\label{s2}
Traditional in-vehicle networks are mostly based on bus technology that can not keep up with the growing communication demands of the self-driving car. In fact, they cannot meet the in-vehicle network requirements for high bandwidth, reliability and real-time communication expectations. Automotive Ethernet, a novel in-vehicle network communication technology, is implemented to ensure an appropriate level of quality of service (QoS) which is essential for time-critical automotive applications.

\begin{figure}[h]
\centering
\includegraphics[scale=0.3]{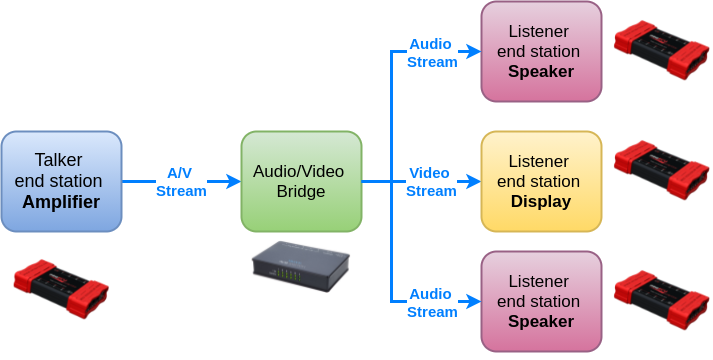}
\caption{Typical AVB Ethernet Local Area Network (LAN).}
\label{fig:avb_network}
\centering
\end{figure}

Audio Video Bridging (AVB) over Ethernet, a set of technical standards, provides improved synchronization, low-latency, and reliability for switched Ethernet networks between multimedia devices. Recently, a lot of automotive products such end-nodes device (i.e.,speakers, cameras, digital signal processors) and network hub (i.e., AV Bridges) support it. In fact, end-nodes can be a talker, a listener or both. A talker is the transmitter of a data stream or the source of the AVB stream and a listener is the receiver or the destination of the AVB stream. These end-nodes are connected by an AVB Bridge which acts as a switch that receives time-critical data from the AVB talker and forwards it to the AVB listener. This interconnection between these three components, as presented in Fig.\ref{fig:avb_network}, is called AVB Ethernet Local Area Network (LAN). 

\begin{figure}[h]
\centering
\includegraphics[scale=0.28]{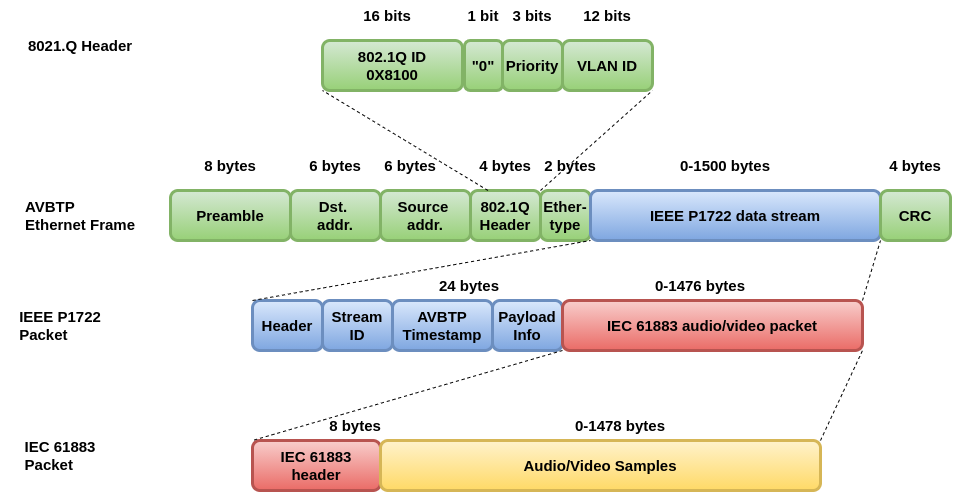}
\caption{IEEE 1722 packet format. Source:  \cite{auto_book}}
\label{fig:avtp_packet}
\centering
\end{figure}

As previously mentioned, AVB has diverse sub-standards to support time-critical in-vehicle applications such as IEE 802.1 Qav, IEEE 802.1 Qat, IEEE 802.1 AS and IEEE 1722. Due to the lack of publicly available datasets which covers attacks on diverse AVB protocols, we are only considering published ones which are composed of captured automotive cyberattacks on IEEE 1722, a stream transmission protocol in charge of transporting control data and audio and video streams. Unfortunately, datasets which cover attacks on other AVB protocols aren't publicly available. As depicted in Fig. \ref{fig:avtp_packet}, the IEEE 1722 packet and its content are sent through an Ethernet frame. The IEEE 802.1Q header is also included in the Ethernet packet. Furthermore, the priority information encapsulated within is critical for the functioning of AVB QoS concept. Moreover, only AVB listener members that share the same AVB talker's VLAN tag can receive the audio/video stream. In the case of The IEEE 1722, the ethertype field's hexadecimal value is 0X22F0. 

In terms of IEEE 1722 streaming packets, the header, the stream ID, the "Presentation time," payload information, and the payload itself are all included therein. The data type of the A/V stream is specified in the header which also includes its sequence number needed by AVB listeners to detect missing packets. The MAC address of the talker is used to produce the stream ID, which identifies a single data stream. The format of the data within the payload is directly related to the field of payload information. The AVBTP timestamp is a time presentation which specifies when a received packet should be delivered to the AVB listener application \cite{auto_book}.  

We will provide in Section \ref{dataset_description} the threat model and the corresponding replay attacks on AVTP protocol, created by Jeong et al. \cite{avtp}, and list also the relevant AVTP features to be leveraged for anomaly detection. 

\section{AVTP Dataset Description}
\label{s3}
\label{dataset_description}
Given a sequence of AVTP packets, we aim to detect whether this sequence is normal or anomalous, i.e., an AVTP sequence is anomalous if it contains at least one abnormal (i.e., injected/out of order/replayed) packets. Hence, we have used the "Automotive Ethernet Intrusion Dataset" dataset \cite{avtp_dataset} created by  Jeong et al. \cite{avtp} and which contains benign and malicious AVTP packet captures from their physical Automotive Ethernet testbed. 

\begin{table}[h]
\begin{center}
\caption{Automotive Ethernet Intrusion Dataset\label{table:avtp_dataset}}
\begin{tabular}{|l | l | l| l|} 
\hline
\multirow{2}{*}{\textbf{Dataset}} & \textbf{\# Normal} & \textbf{\# Abnormal} & \textbf{Size} \\
& \textbf{ packets} & \textbf{packets} & \textbf{(MB)} \\
\hline
$\mathcal{R}$ & 0  & 36 & 0.0164\\
\hline
$\mathcal{D}_{normal}$ & 139,440 & N/A & 63.3\\
\hline
$\mathcal{D}_{injected}^{1}$  & 139,440 &  65,988 & 93.3\\
\hline
$\mathcal{D}_{injected}^{2}$ & 307,020  & 130,906 & 198.8\\
\hline
\end{tabular}
\end{center}
\end{table}

The datasets are recorded in the PCAP file format and, therefore, are viewed using prevalent programming libraries and packet analyzers (such as Wireshark). In fact, the dataset contains four benign (attack-free) packet captures and four malicious ones collected in different environments. The malicious packet captures represent \textbf{replay cyberattack}. In fact, they contain \textbf{message injection} of arbitrary stream AVTP data units (AVTPDUs) into the IVN since the attacker's goal is to output a single video frame, at a terminal application connected to the AVB listener, by injecting previously generated AVTPDUs during a certain period. For our experiment, we have only considered the AVTP packets collected indoor, presented in Table \ref{table:avtp_dataset}. We refer readers to \cite{avtp} for further information. 

 In order to represent AVTP sequences, we use the \textbf{Feature-based Sliding Window (FSW)}\cite{fsw1} to group packets which belong to an AVTP dataset into subsequences with fixed window size $\mathcal{w}$, where $\mathcal{w}$ $\in \{8,16,24,32,40\}$ and the slide size is 1. Hence, each sequence of ordered packets is defined as $S=\{ \mathbf{p}_{1},..., \mathbf{p}_{t},...,\mathbf{p}_{T}\}$, where $\textbf{p}_{t}$ $\in$ $\mathcal{D}$ indicates a transmitted AVTP packet at time $t$, and $\mathcal{D}$ indicates the original AVTP Dataset. Each packet $\mathbf{p}_{t}$ in the AVTP dataset has 438 bytes/features, each of which has a integer value between 0 and 255,  where $\mathbf{p}_{t} \in \mathbb{ Z}^{58}$ (since the most suitable number of bytes used to detect anomalies is the first 58 bytes of each AVTP packet,\cite{avtp}). Hence, to achieve our previously mentioned goal, we train our model using the dataset
$\mathcal{D}_{training}$ 
composed of normal AVTP sequences with each packet. The normal sequences are extracted from dataset $\mathcal{D}_{normal}$, depicted in Table \ref{table:avtp_dataset}. However, when testing, we have preprocessed packets into sequences from both datasets $\mathcal{D}_{injected}^{1}$ and $\mathcal{D}_{injected}^{2}$, and which contain replayed packets from dataset $\mathcal{R}$ . Moreover, We label each AVTP sequence using the following criteria: 
\begin{equation*}
 Y = \begin{cases} \mbox{0 (normal)} & \textup{if }  ( \mathbf{p}_{t} \in S ) \& ( \mathbf{p}_{t}  \notin \mathcal{R} ) , \forall t \in \{1, .., w\} \nonumber \\ 
                      \mbox{1 (abnormal)} & \textup{otherwise}
                \end{cases} 
\end{equation*}
where $Y$ is an AVTP sequence's label, and $\mathcal{R}$ is a set of replayed AVTP packets, collected during a legitimate AVTP media transmission. 

It's worth noting that we do not follow \cite{avtp} labeling criteria. In fact, \cite{avtp} aim to detect packets which are replayed. However, we aim to detect whether a sequence contains one or several injected packets. Notably, using this labeling criteria, more suitable for self-supervised learning, our model can be further used for the detection of cyberattacks different than replay attacks in future work and which are detected by inspecting a series of ordered packets. 

Moreover, we have reshaped our dataset to suit different types of models. Hence, since convolutional autoencoders, presented in Section \ref{s4}, deal with image samples we had to reshape each sequence $S$ into 2D images using the following mapping
\begin{equation}
\label{2D_seq}
Im(S_{k}) = 
\begin{pmatrix}
a_{k,1} & a_{k,2} & \cdots & a_{k,58} \\
a_{k+1,1} & a_{k+1,2} & \cdots & a_{k+1,58} \\
\vdots  & \vdots  & \ddots & \vdots  \\
a_{k+w,1} & a_{k+w,2} & \cdots & a_{k+w,58} 
\end{pmatrix}
\end{equation}
where $Im(S_{k})$ denotes the  $k$-th reshaped sequence of $S$ (corresponding to training sample $k \in \{1,..., N \}$), $w$ is the total sequence length, $a_{m,n}$ is an AVTP packet feature (byte) such that $ 0 \leq a_{m,n} \leq 255$, $ k \leq m \leq k+w$,  and $ 1 \leq n \leq 58 $.  
Hence, an AVTP dataset, represented as $\mathcal{D}=\{Im(S_{k})\}_{k=1}^N$, is ready to be fed into our proposed CAE model. It's worth noting, that when fed into LSTM models, we use $\mathcal{D}=\{S_{k}\}_{k=1}^N$, where $S_k$ is defined as the kth AVTP sequence.

\section{Unsupervised Intrusion Detection Systems}
\label{s4}
Intrusion Detection Systems (IDSs) are considered as an efficient tool to guarantee the confidentiality, integrity and availability of network data. In fact, network intrusions can be detected and identified by comparing their attack signatures to a dataset which contains a pre-defined list of cyberattack patterns. This approach is called signature-based intrusion detection. However, a regular updating of signature databases is not practicable because of the constant evolution of innovative attack tactics. An alternative solution could be the adoption of anomaly-based IDSs which find pattern in the data that deviates from other observations and indicates the presence of malicious activities in the network traffic. In this work, we will compare the performance of deep learning based intrusion detecton systems especially autoencoders based models with state-of-the-art machine learning models.

\subsection{Deep Learning-based IDS }
Deep learning techniques are increasingly used to address the development of complex anomaly detection based IDSs \cite{dl_survey}. One of the most commonly studied feature learning techniques is the use of autoencoder (AE) neural networks which can be used to detect anomalies in high-dimensional data and for different data types including images/videos, sequence data and graph data. 

The autoencoder AE, introduced by Rumelhart et al. \cite{AE}, seeks to learn a low-dimensional feature representation space suitable for reconstructing the provided data instances. During the encoding process, the encoder maps the original data onto low-dimensional feature space, while the decoder tries to retrieve the original data from the projected low-dimensional space. Reconstruction loss functions are used to learn the parameters of the encoder and decoder networks. Its reconstruction error value must be minimized during the training of normal instances and therefore used during testing as an anomaly score. In other words, compared to the typical data reconstruction error, anomalies that differ from the majority of the data have a large data reconstruction error. The following equations govern the behavior of an AE: 

\begin{equation}
\label{z}
    \textbf{z}=\phi_{e}(\textbf{x};\Theta_{e}), 
    \textbf{x} \in \mathbb{R}^d.
\end{equation}
\begin{equation}
\label{xhat}
    \hat{\textbf{x}}=\phi_{d}(\textbf{z};\Theta_{d}),
    \textbf{z} \in \mathbb{R}^m, \hat{\textbf{x}} \in \mathbb{R}^d~, ~ m < d.
\end{equation}
\begin{equation}
\label{min}
    \{\Theta_{e}^{*},\Theta_{d}^{*}\}= \underset{\Theta_{e},\Theta_{d}}{argmin}~(s_{x})
\end{equation}
\begin{equation}
\label{z}
\textup{where} , ~~  s_{x}= \frac{1}{N} \sum_{k=1}^{N} \Vert \textbf{x}_{k} - \hat{\textbf{x}_{k}}\Vert_2^{2}
\end{equation}
where $\textbf{x}_k$ is d-dimensional input for sample $k \in \{1,.., N \}$, N is the number of samples, $\phi_{e}$ is the encoding network with the parameters $\Theta_{e}$, $\textbf{z}$ is an m-dimensional encoding representation of $\textbf{x}$, $\phi_{d}$ is the decoding network with the parameters $\Theta_{d}$, $\hat{\textbf{x}_k}$ is a d-dimensional reconstruction (output of AE for sample $k \in \{1,.., N \}$), $\Theta_{e}^{*}$ and $\Theta_{d}^{*}$ are the optimum values for the encoding and decoding parameters obtained after training the AE through backpropagation, and $s_{x}$ is the mean squared reconstruction error, \cite{dl_survey}.

In fact, only data with normal instances are used to train the AE. Hence, since normal samples in the test dataset have likewise normal profile of training samples, the corresponding reconstruction error is alike. However, compared to the anomalous testing samples, the reconstruction error is much higher. As a result, we can simply classify samples by defining a threshold for reconstruction error: 
\begin{equation}
 c(\textbf{x}) = \begin{cases} \mbox{0 (normal)} & s_{x} < \beta \\ 
                               \mbox{1 (abnormal)} & s_{x} > \beta 
                \end{cases} 
\end{equation}
where c(\textbf{x}) is the classification function for input sample \textbf{x} and $\beta$ is the pre-defined anomaly detection threshold. 

Through our work, we will investigate the performance of two types of autoencoders: Convolutional based autoencoder (CAE), and Long Short Term Memory based autoencoder (LSTMAE). To implement these models, we leverage the Python deep learning framework Pytorch \cite{pytorch}. We train and evaluate them on NVIDIA® Tesla® V100S with 32 GB HBM2 memory. After hyperparameter tuning, we use the commonly chosen hyperparameters depicted in Table \ref{table:model_conf}. 

\begin{table}
\begin{center}
\caption{ AE Models Configuration \label{table:model_conf}}
\begin{tabular}[h]{|l|l |} 
\hline
\textbf{Parameter} & \textbf{Value}\\
\hline
Learning Rate & 0.0001 \\
\hline
Optimizer & Adam \\
\hline
Batch Size & 16 \\
\hline
Early stopping & Yes \\
\hline
\end{tabular}
\end{center}
\end{table}
\begin{table*}
\begin{center}
\caption{CAE's Model Architecture \label{table:cae_dimensions}}
\begin{tabular}[h]{|l|l | l |l|l|l|l|l|} 
\hline
\textbf{Block} & \textbf{Layer} & \textbf{Dimensions} & \textbf{Act. Function} & \textbf{Filter Size}&\textbf{Stride}&\textbf{Padding}&\textbf{Output Padding}\\
\hline
-&Input & (1,$w$,58) & - & - & - & - & -\\
\hline
\multirow{4}{*}{Encoder}&Conv1 & (32,$w$/2,29) & ReLU & (3,3) & (1,1) & (2,2) &-\\
\cline{2-8}
&Conv2 & (64,$w$/4,15) & ReLU & (3,3)& (1,1) & (2,2)&-\\
\cline{2-8}
&Conv3 & (128,$w$/8,8) & ReLU & (3,3) &(1,1) &(2,2)&-\\
\cline{2-8}
&Flatten & (128*$w$) & ReLU & (3,3) &- &-&-\\
\hline
Embedding & Linear & (128*$w$/2) & ReLU & - &- &-&-\\
\hline
\multirow{5}{*}{Decoder}& Unflatten & (128,$w$/8,8) & ReLU & - &- & -&-\\
\cline{2-8}
&Deconv1 & (64,$w$/4,15) & ReLU &(3,3) & (1,1)&(2,2)& (1,0)\\
\cline{2-8}
&Deconv2 &  (32,$w$/2,29) & ReLU & (3,3)&(1,1) &(2,2)& (1,0)\\
\cline{2-8}
&Deconv3 & (1,$w$,58) & ReLU & (3,3)&(1,1) &(2,2)& (1,1)\\
\hline
\end{tabular}
\end{center}
\end{table*}
\subsubsection{Convolutional based Autoencoder (CAE)}
Researchers have been widely using CAE for the anomaly detection of concrete defects \cite{Chow}, in automated video surveillance \cite{Ribeiro}, on system logs \cite{Cui}, and on network application protocols such as HTTP \cite{http}. In fact, CAE is composed of convolutional and deconvolutional layers leveraged in the encoder and decoder parts, respectively. In order to use such architecture, input samples must be reshaped into images. 

The following equation governs the behavior of both convolutional and deconvolutional layers: 
\begin{equation}
    h^{[l+1]}_{k}=\textbf{f}(\sum_{j \in J} x^{[l]}_{j} \circ w^{[l]}_{k} + b_{k}) ,
\end{equation}
where $h^{[l+1]}_{k}$ is the latent representation of $k-th$ feature map in layer {l+1}, \textbf{f} is a non-linear activation function, $x^{[l]}_{j}$ is the $j-th$ feature map of the output layer in layer {l}, $w^{[l]}_{k}$ is the $k-th$ filter weight for the layer {l} and $b_{k}$ is the bias parameter, and $\circ$ represents a 2D convolution operation. 

For different sequence length $w$, we have developed different CAE architectures. Our CAE architecture, depicted in Table. \ref{table:cae_dimensions}, is composed of three convolutional layers on the encoder side, flatten and unflatten layers, one embedding layer, and three deconvolutional layers on the decoder side. For the encoding module, we firstly stack three convolution layers with 36, 64, 128 feature maps, respectively. We have chosen 3x3 kernel sizes for the different convolutional layers, and set the padding and the stride is set to (1,1) and (2,2), respectively. Then we flatten the output of the encoder and feed it to a dense layer that represents the latent space and which is composed of 64 * $w$ neurons (chosen after tuning the correspondent number of neurons). The embedded vector is then unflattened and fed into the decoder. As for the decoding module, we flip the architecture of the encoder, i.e., the corresponding feature maps from bottom to up are 128, 64,32 and 1, and the kernel sizes are 3×3. We set the stride to (1,1), the paddinng to (2,2) and the output padding to (1,0), (1,0) and (1,1) for the three deconvolutional layers.

\begin{table}[h]
\begin{center}
\caption{LSTMAE's Model Architecture \label{table:lstmae_dimensions}}
\begin{tabular}{|l|l | l |l|} 
\hline
\textbf{Block} & \textbf{Layer} & \textbf{Output} & \textbf{Activation} \\
               &                & \textbf{Dimensions} & \textbf{Function} \\
\hline
-&Input & ($w$,58) & -\\
\hline
\multirow{1}{*}{Encoder}&LSTM1 & ($w$,20) & ReLU \\
\hline
Embedding & LSTM2 & (1,10) & ReLU\\
\hline
\multirow{3}{*}{Decoder}& Repeat & ($w$,10) & ReLU\\
\cline{2-4}
&LSTM1 & ($w$,10) & ReLU\\
\cline{2-4}
&LSTM2& ($w$,20)  & ReLU\\
\cline{2-4}
&Linear& ($w$,58)  & -\\
\hline
\end{tabular}
\end{center}
\end{table}

\subsubsection{Long Short Term Memory based Autoencoder}
Long short-term memory based Autoencoder (LSTM-AE), widely used for anomaly detection \cite{lstm1}\cite{lstm2}, are an implementation of autoencoders that uses LSTM as learning layers both in encoder and decoder components. In fact, LSTM networks are a variant the traditional Recurrent Neural Network (RNN) widely used for sequence modeling \cite{lstm3}.

Each LSTM unit consists of three gate structures: an input gate, a forget gate, and an output gate. The input and output gates regulate the memory cell's input and output activation, respectively, whilst the forget gate updates the cell's state. The following equations govern the behavior of an LSTM unit: 
\begin{equation}
 f_{t}=\sigma(W_{xf} \cdot x_{t} + W_{hf} \cdot  h_{t-1} + b_{f})   
\end{equation}
\begin{equation}
 i_{t}=\sigma(W_{xi} \cdot x_{t} + W_{hi} \cdot  h_{t-1} + b_{i})   
\end{equation}
\begin{equation}
 \tilde{C}_{t}=\tanh(W_{xa} \cdot x_{t} + W_{ha} \cdot  h_{t-1} + b_{a})   
\end{equation}
\begin{equation}
 o_{t}=\sigma(W_{xo} \cdot x_{t} + W_{ho} \cdot  h_{t-1} + b_{o})   
\end{equation}
\begin{equation}
C_{t}=f_{t} \otimes C_{t-1} + i_{t} \otimes \tilde{C}_{t}
\end{equation}
\begin{equation}
h_{t}=o_{t} \otimes tanh(C_{t})
\end{equation}
where $h_{t-1}$ and $C_{t-1}$ are output and cell state at the previous moment, respectively, $x_{t}$ represents the current input, f represents the forget gate, $f_{t}$ is a forget control signal which determines if the prior unit's state $C_{t-1}$ should be reserved, $f_{t} \otimes C_{t-1} $ represents the information retained at the previous moment, i represents the input gate,  $\tilde{C}_{t}$ is considered as the candidate cell state at time t, $i_{t}$ represents the control signal for $\tilde{C}_{t}$, $h_{t}$ is regarded as the final output, $o_{t}$ represents the output control signal. Moreover, $\{ W_{x,i}~,~ W_{x,f} ~,~ W_{x,a}~,~W_{x,o}  \}$ represents the  $\{ $ input, forget, active, output $\} $-layer connection matrices (all of which to be learned), and $\{ W_{h,i}~,~ W_{h,f} ~,~ W_{h,a}~,~W_{h,o}  \}$ indicate the $\{ $ input, forget, active, output $\} $-hidden layer recurrent connection matrices (all to be optimized), $\sigma$ is the sigmoid activation function and $\otimes$ represents element-wise (Hadamard) product.

In fact, for each sequence length $w$, we create a different LSTMAE. As shown in Table \ref{table:lstmae_dimensions}, for the LSTM based encoding module, we firstly stack two LSTM layers which output an embedding vector of size 10 (chosen after tuning). Then we repeat the embedding vector $w$ times, and feed it into the decoder. As for the decoding module, we flip the architecture of encoder, i.e. the repeated vector passes through two LSTM layers with number of features 10 and 20 respectively and a dense layer, to be finally reconstructed.
\subsubsection{Anomaly Detection using AE models}
As previously mentioned, we will classify AVTP sequence samples by defining a threshold $\beta$. Hence, after training our AE models for each window size $w$, we vary $\beta$ between $\mu - \alpha_{min} \sigma$ and $\mu + \alpha_{max} \sigma$ where $\mu$ is the mean reconstruction error of normal samples used for training, $\sigma$ is the standard deviation of normal samples' reconstruction errors, $\alpha \in \{-2,2\}$ with a step size $\delta=0.5$, $\alpha_{max}=max(\alpha)$ and $\alpha_{min}=min(\alpha)$ to select the best threshold.
\begin{figure*}
\centering
\begin{tikzpicture}
\begin{groupplot}[
       group style={
       group name=plot,
       group size=1 by 3,
       xlabels at=edge bottom,
       ylabels at=edge left,
       horizontal sep=0pt,
       vertical sep=15pt,
       /pgf/bar width=9.5pt},
major x tick style=transparent,
ybar= \pgflinewidth,
ymax=1,
x axis line style={opacity=0},
x tick label style={rotate=15, anchor=center},
xticklabel style={yshift=-2mm,xshift={ifthenelse(\ticknum==2,0,0)}}, 
xtick=data,
xticklabels={},
ymajorgrids=true,
grid style=dotted,
nodes near coords,
scale only axis,
point meta=explicit symbolic,
enlarge x limits = {abs=1},
cycle list={
  draw=blue,thick,fill=blue,fill opacity=0.6,nodes near coords style={blue!60}\\
  draw=orange,thick,fill=orange,fill opacity=0.6,nodes near coords style={orange}\\
  draw=green,thick,fill=green,fill opacity=0.6,nodes near coords style={green}\\
  draw=red,thick,fill=red,fill opacity=0.6,nodes near coords style={red}\\
  draw=black,thick,fill=black,fill opacity=0.6,nodes near coords style={black}\\
  draw=purple,thick,fill=purple,fill opacity=0.6,nodes near coords style={purple}\\
  },
height=0.1\textwidth,
]

\nextgroupplot[
     width=\textwidth,
     ytick pos=left,
     ylabel={F1-score},
     legend style={at={(0.5,+1.5)},legend
        columns=6,fill=none,draw=black,anchor=center,align=center}
      ]
\addplot  coordinates {
  (1,0.7051)
  (2,0.8635)[0.86]
  (3,0.9388)[0.94]
  (4,0.9711)[0.97]
  (5,0.9825)[0.98]
  };

\addplot  coordinates {
  (1,0.7635)[0.76]
  (2,0.7799)
  (3,0.8095)
  (4,0.8297)
  (5,0.8568)
};
  
  \addplot  coordinates {
  (1,0.4809)
  (2,0.5143)
  (3,0.5431)
  (4,0.5667)
  (5,0.5854)
};
  
   \addplot  coordinates {
  (1,0.0736)
  (2,0.0938)
  (3,0.09)
  (4,0.0781)
  (5,0.043)
};
  
   \addplot  coordinates {
  (1,0.2089)
  (2,0.2074)
  (3,0.1917)
  (4,0.1852)
  (5,0.1326)
  };
\legend{CAE,LSTMAE,OCSVM,LOF,IF}

\nextgroupplot[
     width=\textwidth,
     ylabel={Recall},
     ytick pos=left,
      ]
\addplot  coordinates {
  (1,0.5508)
  (2,0.7824)
  (3,0.9081)
  (4,0.9635)
  (5,0.9806)
  };

\addplot  coordinates {
  (1,0.833)
  (2,0.9089)
  (3,0.9006)
  (4,0.9534)
  (5,0.95)
};

  \addplot  coordinates {
  (1,0.5031)
  (2,0.4988)
  (3,0.4991)
  (4,0.4989)
  (5,0.4982)
};
  
   \addplot  coordinates {
  (1,0.0397)
  (2,0.0497)
  (3,0.0472)
  (4,0.0407)
  (5,0.0220)
};
  
   \addplot  coordinates {
  (1,0.1608)
  (2,0.1508)
  (3,0.1251)
  (4,0.1178)
  (5,0.0785)
  };

\nextgroupplot[
     width=\textwidth,
     xticklabels={8,16,24,32,40},
     xlabel={$w$},
     ylabel={Precision},
     ytick pos=left,
      ]
\addplot  coordinates {
  (1,0.9794)
  (2,0.9634)
  (3,0.9718)
  (4,0.9789)
  (5,0.9844)
  };

\addplot  coordinates {
  (1,0.7047)
  (2,0.683)
  (3,0.7352)
  (4,0.7344)
  (5,0.7801)
};
  
  \addplot  coordinates {
  (1,0.4605)
  (2,0.5307)
  (3,0.5954)
  (4,0.6560)
  (5,0.7096)
};
  
   \addplot  coordinates {
  (1,0.5022)
  (2,0.8337)
  (3,0.9381)
  (4,0.9760)
  (5,0.9847)
};
  
   \addplot  coordinates {
  (1,0.2980)
  (2,0.3325)
  (3,0.4091)
  (4,0.4322)
  (5,0.4269)
  };

\end{groupplot}
\end{tikzpicture}
\caption{Comparision of different unsupervised machine learning anomaly detection performance under different window sizes $w$ on $\mathcal{D}^{1}_{injected}$ }
\label{fig:comparision_d1}
\end{figure*}

\begin{figure*}
\centering
\begin{tikzpicture}
\begin{groupplot}[
       group style={
       group name=plot,
       group size=1 by 3,
       xlabels at=edge bottom,
       ylabels at=edge left,
       horizontal sep=0pt,
       vertical sep=15pt,
       /pgf/bar width=9.5pt},
major x tick style=transparent,
ybar= \pgflinewidth,
ymax=1,
x axis line style={opacity=0},
x tick label style={rotate=15, anchor=center},
xticklabel style={yshift=-2mm,xshift={ifthenelse(\ticknum==2,0,0)}}, 
xtick=data,
xticklabels={},
ymajorgrids=true,
grid style=dotted,
nodes near coords,
scale only axis,
point meta=explicit symbolic,
enlarge x limits = {abs=1},
cycle list={
  draw=blue,thick,fill=blue,fill opacity=0.6,nodes near coords style={blue!60}\\
  draw=orange,thick,fill=orange,fill opacity=0.6,nodes near coords style={orange}\\
  draw=green,thick,fill=green,fill opacity=0.6,nodes near coords style={green}\\
  draw=red,thick,fill=red,fill opacity=0.6,nodes near coords style={red}\\
  draw=black,thick,fill=black,fill opacity=0.6,nodes near coords style={black}\\
  draw=purple,thick,fill=purple,fill opacity=0.6,nodes near coords style={purple}\\
  },
height=0.1\textwidth,
]

\nextgroupplot[
     width=\textwidth,
     ytick pos=left,
     ylabel={F1-score},
     legend style={at={(0.5,+1.5)},legend
        columns=6,fill=none,draw=black,anchor=center,align=center}
      ]
\addplot  coordinates {
  (1,0.7288)
  (2,0.8823)[0.88]
  (3,0.9507)[0.95]
  (4,0.977)[0.98]
  (5,0.9857)[0.98]
  };

\addplot  coordinates {
  (1,0.7425)[0.74]
  (2,0.7639)
  (3,0.7946)
  (4,0.8159)
  (5,0.8331)
};

  \addplot  coordinates {
  (1,0.4816)
  (2,0.5184)
  (3,0.5448)
  (4,0.5662)
  (5,0.5855)
};
  
   \addplot  coordinates {
  (1,0.058)
  (2,0.0834)
  (3,0.0727)
  (4,0.0629)
  (5,0.0375)
};
  
   \addplot  coordinates {
  (1,0.2339)
  (2,0.2062)
  (3,0.2008)
  (4,0.1563)
  (5,0.1655)
  };
\legend{CAE,LSTMAE,OCSVM,LOF,IF}

\nextgroupplot[
     width=\textwidth,
     ylabel={Recall},
     ytick pos=left,
      ]
\addplot  coordinates {
  (1,0.5779)
  (2,0.8008)
  (3,0.9186)
  (4,0.967)
  (5,0.9832)
  };

\addplot  coordinates {
  (1,0.8156)
  (2,0.796)
  (3,0.8925)
  (4,0.9487)
  (5,0.8634)
};

  \addplot  coordinates {
  (1,0.5120)
  (2,0.5116)
  (3,0.5083)
  (4,0.5041)
  (5,0.5017)
};
  
   \addplot  coordinates {
  (1,0.031)
  (2,0.0438)
  (3,0.0378)
  (4,0.0325)
  (5,0.0191)
};
  
   \addplot  coordinates {
  (1,0.1779)
  (2,0.1388)
  (3,0.1323)
  (4,0.0944)
  (5,0.099)
  };

\nextgroupplot[
     width=\textwidth,
     xlabel={$w$},
     xticklabels={8, 16, 24, 32, 40},
     ylabel={Precision},
     ytick pos=left,
      ]
\addplot  coordinates {
  (1,0.9863)
  (2,0.9824)
  (3,0.9851)
  (4,0.9871)
  (5,0.9882)
  };

\addplot  coordinates {
  (1,0.6814)
  (2,0.7343)
  (3,0.716)
  (4,0.7157)
  (5,0.8049)
};

  \addplot  coordinates {
  (1,0.4545)
  (2,0.5255)
  (3,0.5870)
  (4,0.6457)
  (5,0.7029)
};
  
   \addplot  coordinates {
  (1,0.4420)
  (2,0.8726)
  (3,0.9463)
  (4,0.9725)
  (5,0.9757)
};
  
   \addplot  coordinates {
  (1,0.3414)
  (2,0.4005)
  (3,0.4156)
  (4,0.4552)
  (5,0.5037)
  };

\end{groupplot}
\end{tikzpicture}
\caption{Comparison of different unsupervised machine learning anomaly detection performance under different window sizes $w$ on $\mathcal{D}^{2}_{injected}$ }
\label{fig:comparision_d2}
\end{figure*}
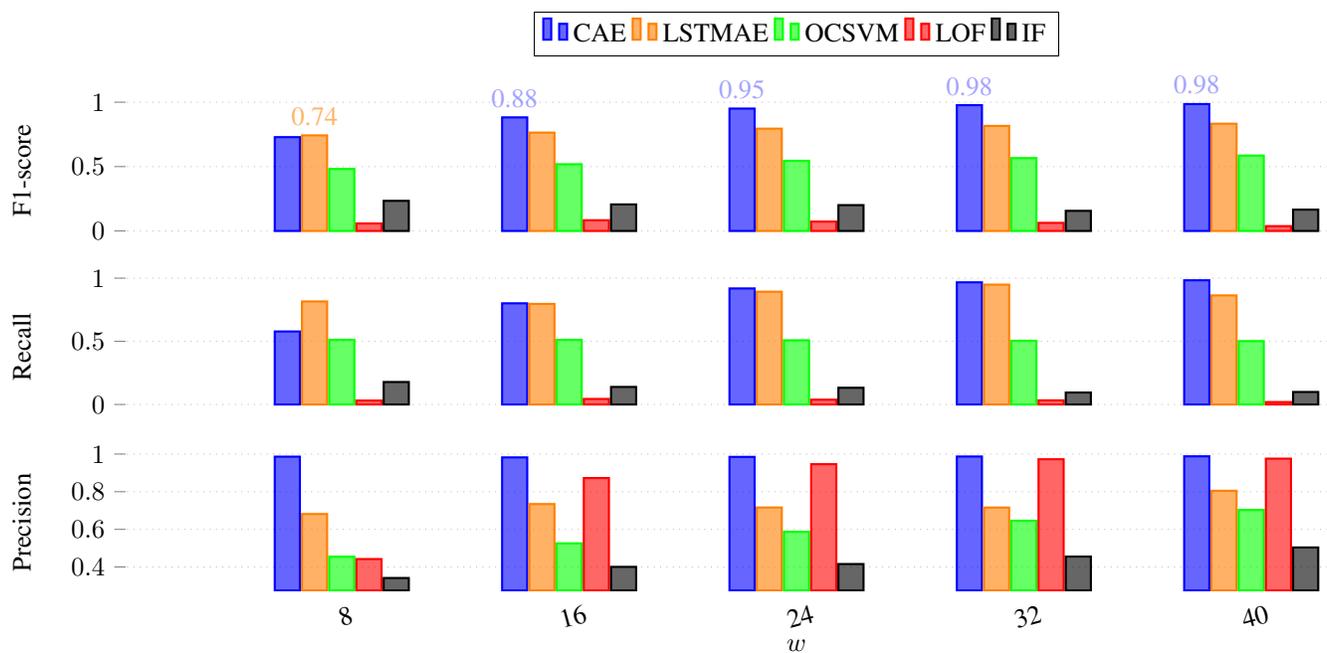
\subsection{Machine Learning-based IDS }
Through our work, we compare the autoencoder based models to state-of-the-art machine learning based anomaly detection algorithms: One-Class SVM (OCSVM), Local Outlier Factor (LOF) and Isolation Forest (IF). We have implemented these algorithms using Scikit-learn python library, and have trained and evaluated them on an 3.3 Ghz AMD EPYC™ 7402. 
\subsubsection{One-Class SVM}
One-Class Support Vector Machine (OCSVM), proposed by Scholkopf et al. \cite{ocsvm}, is an appealing tool for anomaly detection in different fields, such as the detection anomalies in EEG data from epilepsy patients \cite{EEG}, document classification \cite{document_classification} and others. Using OCSVM, data are first mapped into a feature space using an appropriate kernel function and then maximally separated from the origin using a hyperplane. After tuning, we have used the radial basis function (RBF) as the kernel function.
\subsubsection{Local Outlier Factor}
Local Outlier Factor(LOF), originally proposed by Breunig et al. \cite{lof}, is a density-based outlier detection algorithm that finds outliers by calculating the local deviation of a given data point \cite{lof1} \cite{lof2}. In fact, LOF value of normal data is approximately equal to 1, while the outlier value is significantly higher than 1. In other words, if a sample is located within a cluster, its localized density is similar to the nearest neighbour. Hence, its value is close to 1. 
\subsubsection{Isolation Forest}
Isolation forest (IF), proposed by Liu at al. \cite{IsolationForest}, detects anomalies using isolation rather than modelling the normal points. In fact, this technique presents a novel approach for isolating anomalies using binary trees, providing a new prospect for a speedier anomaly detector that directly targets abnormalities rather than profiling all regular instances.
\begin{table*}

    \caption{AutoEncoder-based Models' Characteristics \& Computational Resources}
    \label{tab:computation}
    \centering

    \begin{tabular}{|c| c  c  c |c  c  c|}
    \hline
      & \multicolumn{3}{c|}{\textbf{Conv-AE } }& \multicolumn{3}{c|}{\textbf{LSTM-AE} }\\
    \cline{2-7}
    \multirow{3}{*}{\textbf{Window} } &  &  &  & &  & \\ 
      & Inference Time & $\#$Parameters & Model Size & Inference Time& $\#$Parameters & Model Size\\

    & (s) &  & (MB) & (s) &  & (KB)\\
    \hline
   8   & $0.49\pm0.53$ & 1,235,329 & 4.8 & $1.17\pm0.90$ & 12,338 & 52\\
    \hline
   16   & $0.48\pm0.52$ & 4,382,593 & 17 & $1.58\pm0.96$ & 12,338 & 52 \\
    \hline
   24  & $0.38\pm0.33$ & 9,627,009 & 37 & $1.91\pm0.95$ & 12,338 & 52 \\
    \hline
   32   & $0.43\pm0.28$ & 16,968,577 & 65 & $2.31\pm1.02$ & 12,338 &  52\\
    \hline
   40  & $0.45\pm0.14$ & 26,407,297 & 101 & $2.70\pm0.99$ & 12,338 & 52 \\
    \hline

    \end{tabular}
\end{table*}
\section{Evaluation metrics}
\label{s5}
For measuring the performance of different anomaly-based IDS, we use the F1-score metric, a weighted average result of both metrics precision and recall and which is specifically used when the dataset is imbalanced. The model has a large predictive power if the F1-score is near 1.0.

Precision is the ratio of correctly classified predicted abnormal observations of all the observations in the predicted class.
\begin{equation}
    Precision = \frac{TP}{TP + FP}
\end{equation}
Recall is the ratio of correctly predicted abnormal observations of all observations in the actual class.
\begin{equation}
    Recall = \frac{TP}{TP + FN}
\end{equation}
Hence, the F1-score is calculated using the following equation:
\begin{equation}
    F1-score= 2 \cdot \frac{Precision \cdot Recall}{Precision + Recall}
\end{equation}

Where: TP= True Positive; FP=False Positive; TN= True Negative; FN=False Negative.

\section{Results}
\label{s6}
Figures \ref{fig:comparision_d1} and \ref{fig:comparision_d2} shows the performance of different anomaly based IDS on both datasets $\mathcal{D}_{injected}^{1}$ and $\mathcal{D}_{injected}^{2}$. As seen, conventional machine learning algorithms such as OCSVM, Isolation Forest, and Local Outlier Factor perform poorly on both datasets when recognizing anomalous AVTP sequences for different sequence length. In fact, these traditional anomaly detection models are inefficient at detecting anomalies in large, high-dimensional datasets since these methods assume small datasets with low numbers of features. Hence, when dealing with a huge input dimensionality, a high proportion of irrelevant features can effectively creates noise in the input data, which masks the true anomalies and engenders poor anomaly detection performance.

To overcome the limitations of these approaches in high-dimensional datasets, the deep learning models CAE and LSTMAE, are considered as a better alternative for anomaly detection. As demonstrated in Figures \ref{fig:comparision_d1} and \ref{fig:comparision_d2}, they significantly outperform the benchmark anomaly detection models and achieve reasonbale F1-scores on both datasets. After tuning the threshold $\beta$ for various sequence length and for different datasets and AE models, the CAE and LSTMAE reached their highest performance when $\beta=\mu+0.5\sigma$. Moreover, the CAE model achieves an overall better performance in terms of F1-score scores than LSTMAE model which indicates that LSTMAE is not able to encode the context information of an AVTP sequence from both the left and right context especially when working on long sequences ($w \ge 16$). Despite the fact that a Bidirectional LSTMAE is commonly used nowadays to represent contextual information, they suffer from the vanishing or exploding gradients. In other words, the model hardly captures the long-term dependency and which is critical for the detection of anomalies in large sequences. When varying the AVTP sequence length between 16 and 40, CAE has outperformed LSTMAE by exploiting significant correlations in a sequence of AVTP packets. The performance of both models proportionally increases when increasing window length on both datasets, since AVTP sequences will contain more injected packets, thus it becomes easier to differentiate between normal and abnormal AVTP sequences. 

We also assess the the performance of the best models, more specifically AE models, by measuring their computational power and their memory requirements. As depicted in Table \ref{tab:computation}, although the CAE model has a bigger number of parameters and a larger model size than LSTMAE, it stays speedier when detecting anomalies in AVTP sequences for different window sizes. Hence, CAE is more suitable for real-time intrusion detection than LSTMAE. Although it has  larger models' size, the CAE models can either be deployed on a cloud server connected to the in-vehicle network or can be embedded inside an ECU with suitable memory characteristics. In the future, we plan to examine the implementation of both ideas.

\section{Limitations}
\label{s7}
Due to the lack of datasets which represents attacks on AVTP protocol and the availability of only one dataset that  solely represents replay attacks, our current comparison between the different deep and machine learning models can't be extrapolated to different types of cyberattacks on AVTP. Thus, our comparison needs further investigation when Automotive Ethernet datasets with diverse types of intrusions are available. In addition, while we validated our solutions for real-time scenarios, we have not yet implemented them on hardware. In the future, we plan to examine their implementation on cloud servers or any ECU connected to the in-vehicle network.

\section{Conclusions and Future Work}
Anomaly detection in in-vehicle network protocols, especially in Automotive Ethernet, is a burgeoning study area. With the development of realistic datasets which represent automotive cyberattacks on this protocol, we are able to develop anomaly-based detection models and to evaluate them. In this paper, we compared the performance of different deep and machine learning algorithms for learning normal Audio Video Transport Protocol (AVTP) communication behavior and thus identify cyberattacks on this protocol. The numerical results show that autoencoder based IDS outperform state-of-the-art traditional machine learning models for different AVTP sequence length. Moreover, convolutional based AE are suitable for real-time intrusion detection. For future work, we aim to perform a similar analysis on a variety of AVB dataset with sophisticated cyberattacks. Furthermore, we aim to study the performance of anomaly detection algorithms on other protocols running on top of Automotive Ethernet such as Scalable Service-Oriented Protocol (SOME/IP) and Diagnosis over IP (DoIP) protocol and which have different network characteristics and vulnerabilities.

\end{document}